\documentclass[10pt,twocolumn,letterpaper]{article}

\usepackage{iccv}
\usepackage{times}
\usepackage{epsfig}
\usepackage{graphicx}
\usepackage{amsmath}
\usepackage{amssymb}

\usepackage{tablefootnote}
\usepackage{subcaption}
\usepackage{color, colortbl}

\definecolor{First}{gray}{0.6}
\definecolor{Second}{gray}{0.7}
\definecolor{Third}{gray}{0.8}
\definecolor{Fourth}{gray}{0.9}

\usepackage{amsthm}
\newtheorem{theorem}{Theorem}
\usepackage[breaklinks=true,bookmarks=false]{hyperref}

\iccvfinalcopy 

\def\iccvPaperID{711} 

\ificcvfinal\fi
\begin{document}

\title{SparseMask: Differentiable Connectivity Learning for Dense Image Prediction}

\author{Huikai Wu\thanks{This paper is based on the results obtained from the author's internship at Preferred Networks, Inc.}~~~~Junge Zhang~~~~Kaiqi Huang\thanks{Also with CAS Center for Excellence in Brain Science and Intelligence Technology.} \\
Institute of Automation, Chinese Academy of Sciences\\
University of Chinese Academy of Sciences\\
{\tt\small\{huikai.wu, jgzhang, kaiqi.huang\}@nlpr.ia.ac.cn}
}

\maketitle
\thispagestyle{empty}

\begin{abstract}
	In this paper, we aim at automatically searching an efficient network architecture for dense image prediction.
	Particularly, we follow the encoder-decoder style and focus on designing a connectivity structure for the decoder.
	To achieve that, we design a densely connected network with learnable connections, named Fully Dense Network, which contains a large set of possible final connectivity structures.
	We then employ gradient descent to search the optimal connectivity from the dense connections.
	The search process is guided by a novel loss function, which pushes the 
		weight of each connection to be binary and the connections to be sparse.
	The discovered connectivity achieves competitive results on two segmentation datasets, while runs more than three times faster and requires less than half parameters compared to the state-of-the-art methods.
	An extensive experiment shows that the discovered connectivity is compatible with various backbones and generalizes well to other dense image prediction tasks.
	Code is available at \url{https://github.com/wuhuikai/SparseMask}.
\end{abstract}

\section{Introduction}
\begin{figure}
\begin{center}
	\begin{subfigure}[t]{0.32\linewidth}
		\includegraphics[height=2\linewidth]{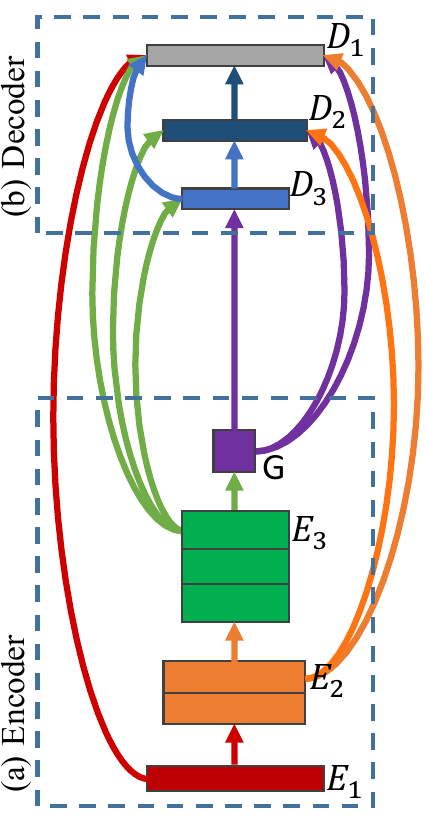}
		\caption{Step 1}
		\label{fig:framework_01}
	\end{subfigure}
	\begin{subfigure}[t]{0.32\linewidth}
		\includegraphics[height=2\linewidth]{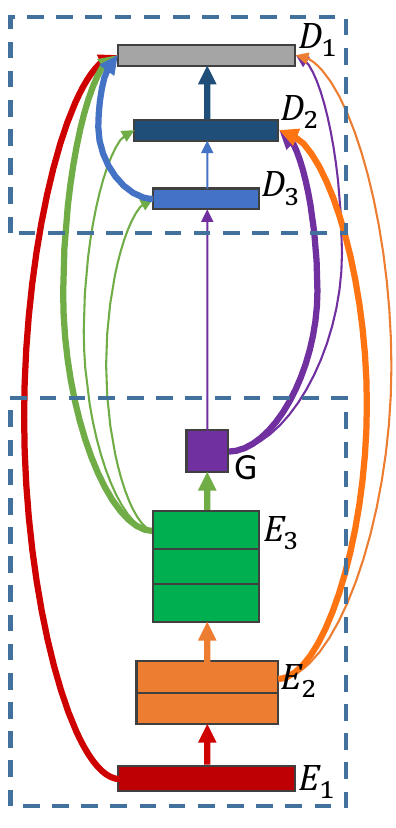}
		\caption{Step 2}
		\label{fig:framework_02}
	\end{subfigure}
	\begin{subfigure}[t]{0.32\linewidth}
		\includegraphics[height=2\linewidth]{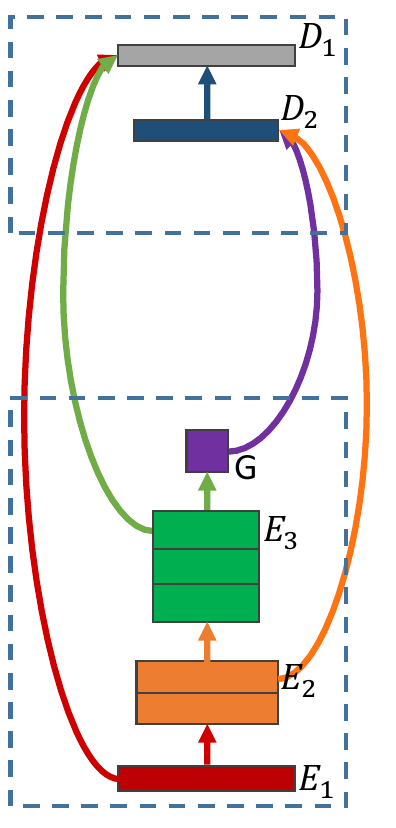}
		\caption{Step 3}
		\label{fig:framework_03}
	\end{subfigure}
\end{center}
	\caption{\textbf{Framework Overview.} (a) Transform the pre-trained image classifier into a Fully Dense Network with learnable connections. (b) Search the optimal connectivity with the proposed sparse loss in a differentiable manner. (c) Drop the useless connections and stages to obtain the final architecture. Best viewed in color.}
\label{fig:framework}
\end{figure}
Dense image prediction is a collection of computer vision tasks that produce a pixel-wise label map for a given image.
Such tasks range from low-level vision to high-level vision, including edge detection~\cite{xie2015holistically}, saliency detection~\cite{hou2017deeply} and semantic segmentation~\cite{long2015fully}.
To address these tasks, Long \etal propose fully convolutional networks (FCNs), which follow an encoder-decoder style~\cite{long2015fully}.
The encoder is transformed from a pre-trained image classifier, while the decoder combines low-level and high-level features of the encoder to generate the final output.
As follows, various methods based on FCNs are proposed, which focus on adjusting the architecture of the decoder manually to achieve a better fusion of multi-level encoder features~\cite{xie2015holistically,ronneberger2015u,lin2017refinenet,hou2017deeply,fu2017stacked,yu2018learning,zhang2018exfuse,peng2017large}.
However, designing the architecture remains a laborious task, which requires lots of expert knowledge and takes ample time.

Inspired by the success of neural architecture search (NAS) in image classification~\cite{zoph2016neural,zoph2017learning,liu2017progressive,liu2018darts}, we aim at automatically designing a decoder for dense image prediction tasks.
However, directly employing the methods from image classification is not sufficient, because
(1) dense image prediction requires producing a pixel-wise label map, while image classification focuses on predicting a class label,
(2) the key to dense image prediction is encoding multi-level features, while image classification aims at extracting global features,
and (3) dense image prediction usually requires a pre-trained image classifier for faster training and better performance, while the network for image classification can be designed and trained from scratch.
Thus, we face two major challenges:
(1) We are required to design a novel search space to combine multi-level features of the pre-trained classifier.
(2) The feature maps are usually in high resolution, bringing in heavy computation complexity and memory footprint.
Thus, the proposed algorithm needs to design a time and memory efficient network.

To solve the first challenge, we propose a densely connected network named Fully Dense Network (FDN) as the search space, which defines a large set of possible final architectures.
FDN follows the encode-decode style, where the encoder is converted from the pre-trained image classifier and the decoder consists of learnable dense connections (Figure~\ref{fig:framework_01}).
To solve the second challenge, we design a novel loss function to search the optimal connectivity in a differentiable way.
The proposed loss forces the weight of each connection to be binary and the connectivity to be sparse (Figure~\ref{fig:framework_02}).
After training, we prune FDN to obtain the final architecture with sparse connectivity (Figure~\ref{fig:framework_03}), which is time and memory efficient.
The three steps above (Figure~\ref{fig:framework}) form the proposed method named SparseMask, which can automatically search an efficient decoder for dense image prediction tasks.
Particularly, the proposed method focuses on designing the connectivity structure to achieve a better fusion of low-level and high-level features.

To validate the effectiveness of our method, we take a comprehensive ablation study on the Pascal VOC 2012 benchmark~\cite{everingham2015pascal} with MobileNet-V2~\cite{sandler2018mobilenetv2} as the backbone.
Results show that our method discovers an architecture that outperforms the baseline methods by a large margin within 18 GPU-hours.
We then transfer the architecture to other backbones, datasets and tasks.
Experiments show that the discovered connectivity has a good generalization ability, which achieves competitive performance, runs more than three times faster, and has less than half parameters.

In summary, we propose a novel method that automatically designs an efficient connectivity structure for the decoder of dense image prediction tasks in a differentiable way.
Our contributions are three-folds,
(1) we propose Fully Dense Network to define the search space,
(2) we introduce a novel loss function to force the dense connections to be sparse,
and (3) we conduct comprehensive experiments to validate the effectiveness of our method as well as the generalization ability of the discovered connectivity.
\section{Related Work}
\subsection{Architecture Search}
Our work is motivated by differentiable architecture search~\cite{saxena2016convolutional,veniat2017learning,liu2018darts,ahmed2018maskconnect}, which is based on the continuous relaxation of the architecture representation, allowing efficient search with gradient descent.
\cite{saxena2016convolutional,veniat2017learning} propose a grid-like network as the search space, while \cite{liu2018darts} relax the search space to be continuous and search the space by solving a bilevel optimization problem.
Other works in architecture search employ reinforcement learning~\cite{baker2016designing,zoph2016neural}, evolutionary algorithms~\cite{real2017large,xie2017genetic,liu2017hierarchical}, and sequential model-based optimization~\cite{negrinho2017deeparchitect,liu2017progressive} to search the discrete space.

To the best of our knowledge, the most related work is MaskConnect~\cite{ahmed2018maskconnect}, which is designed for image classification.
Compared to it, our method is unique in three aspects:
(1) We propose a novel sparse loss, which allows an arbitrary number of input features, resulting in a larger and more flexible search space.
As for MaskConnect, the number of input features is fixed.
(2) We propose an efficient way to concatenate multiple feature maps with different spatial resolutions instead of simply padding and summation.
(3) Following the encoder-decoder style~\cite{ronneberger2015u}, we design a search space for dense image prediction tasks, which differs notably from that of MaskConnect.

As for dense image prediction, \cite{saxena2016convolutional,veniat2017learning,fourure2017residual} propose to embed a large number of architectures in a grid-like network.
However, the searched network has to be trained from scratch.
Differently, our method utilizes a pre-trained image classifier and has a much larger search space.
Our work is also complementary to \cite{chen2018searching}, which constructs a recursive search space and employs random search~\cite{golovin2017google} to discover the best architecture.
Such a method focuses on extracting multi-scale information from the high-level features, while ours aims at the fusion of the low-level and high-level features.
Besides, the search space and search approach are significantly different.
AutoDeepLab~\cite{Liu2019AutoDeepLabHN} is also related to our work.
However, our method focuses on designing the connectivity of the decoder while AutoDeepLab aims at searching the architecture of the encoder.
\subsection{Dense Image Prediction}
Currently, there are two prominent paradigms for dense image prediction.
\cite{xie2015holistically,ronneberger2015u,lin2017refinenet,hou2017deeply,fu2017stacked,yu2018learning,zhang2018exfuse,peng2017large} propose an encoder-decoder network for combining low-level and high-level features, while \cite{chen2014semantic,zheng2015conditional,yu2015multi,chen2017rethinking,zhao2017pyramid,zhang2018context} utilize dilated convolutions to keep the receptive field size and design a multi-scale context module to process high-level features.

In this paper, we follow the encoder-decoder style, of which the key is designing a connectivity structure to fuse low-level and high-level features.
\cite{ronneberger2015u} introduce skip connections to combine the decoder features and the corresponding encoder activations, resulting in a network named U-Net.
Alternatively, \cite{xie2015holistically} aggregate multiple side outputs to generate the final result.
\cite{hou2017deeply} introduce short connections and deep supervision, while \cite{liu2018path} enhance the U-Net architecture with an additional bottom-up path.
Differently, we aim at automatically designing a sparse connectivity structure to fuse the multi-level features more effectively. 
\section{Methods}
In this paper, we aim at automatically transforming a pre-trained image classifier into an efficient fully convolutional network (FCN) for dense image prediction tasks.
Concretely, given a pre-trained image classifier, we follow the three steps as shown in Figure~\ref{fig:framework}.
(1) Transform the classifier into a densely connected network with learnable connections, \ie Fully Dense Network (FDN). See Section~\ref{sec:fdn}.
(2) Employ gradient descent to train the FDN with a novel loss function, which forces the dense connections to be sparse. See Section~\ref{sec:loss}.
(3) Prune the well-trained FDN to obtain the final network architecture. See Section~\ref{sec:prune}.
\subsection{Fully Dense Network}
\label{sec:fdn}
We follow the encoder-decoder style~\cite{badrinarayanan2015segnet,ronneberger2015u} in dense image prediction tasks to design the search space, resulting in a super-network named Fully Dense Network (FDN).
As shown in Figure~\ref{fig:framework_01}, the encoder is a pre-trained image classifier, while the decoder combines multi-level features of the encoder and the decoder with learnable connections.
\subsubsection{The Encoder}
~~~~An image classifier is usually composed of multiple convolution layers, a global average pooling layer~\cite{lin2013network}, and a multi-layer perceptron (MLP).
To transform the classifier into an encoder, we simply drop the MLP and keep the rest of the network unchanged.
As shown in Figure~\ref{fig:framework_01}, the encoder consists of three convolution stages, as well as a global average pooling layer.
Each convolution stage contains multiple convolution blocks, such as residual block~\cite{he2016deep} and inception block~\cite{szegedy2016rethinking}.
Because the features inside a stage usually have the same spatial and channel dimensions, it's reasonable to assume that the feature of the last block contains the most useful information.
We thus restrict the decoder from accessing the other features within a stage.
Concretely, the input features of the decoder is limited to the last feature of each encoder stage and the feature after global average pooling, which are noted as $E_l$ and $G$ respectively.
$l$ is an index ranging from $1$ to $L$, where $L$ is the number of convolution stages.
\subsubsection{The Decoder}
\label{sec:decoder}
~~~~We focus on automatically designing the connectivity structure between the encoder and the decoder, as well as the connections inside the decoder. 
The first problem is to decide the number of stages in the decoder automatically.
To achieve that, we initialize the decoder with a large number of stages and employ the search algorithm to select the most important ones.
Concretely, the decoder is initialized with the same number of stages as the encoder, \ie $L$ stages.
The feature generated by each stage of the decoder is noted as $D_l$, where $l$ is an index ranging from $L$ to $1$, as shown in Figure~\ref{fig:framework_01}.
Additionally, $E_l$ and $D_l$ have the same spatial dimensions by our design.

The second problem is to automatically choose the input features for each decoder stage.
Inspired by DenseNet~\cite{huang2017densely}, we propose to initialize the decoder as a densely connected network.
Many classic architectures in dense image prediction tasks are a subset of the proposed network, such as U-Net~\cite{ronneberger2015u}.
As shown in Figure~\ref{fig:framework_01}, the input features of decoder stage $l$ contains three parts, which are $E_i$ ($i>=l$), $D_i$ ($i>l$) and $G$.
Our method then selects the most important features for each decoder stage automatically.
Notably, the proposed FDN is significantly different from DenseNet in three aspects.
(1) FDN is densely connected in network-level while DenseNet only has block-level dense connections.
(2) The input features of a decoder stage in FDN are from both inside and outside the decoder, following the encoder-decoder style.
(3) The input features of a decoder stage in FDN have different spatial dimensions.

The last problem is to efficiently combine the input features inside each decoder stage.
In MaskConnect, the input features are padded to the largest spatial/channel dimensions and summed into one feature.
The fused feature is then processed by a convolution block.
To make the fusion more flexible, we propose to concatenate all the input features channel-wisely and then apply a convolution block to produce the output.
However, all the features are required to be up-sampled to the same spatial dimensions before concatenating, which has heavy memory footprint and computation complexity.
To reduce memory usage and speed up computation, we show that
(1) concatenating the features and then applying convolution is equal to applying convolution to each feature and then take a summation,
and (2) the order of bilinear upsampling and point-wise convolution is changeable.\footnote{The proofs are shown in the supplementary material.}
Thus, the operations within a decoder stage can be formulated as Equation~\ref{eq:conv},
\begin{equation}
	\begin{aligned}
		D_l &= \sum_{t\in T_l}{w_l^tf_{\uparrow}(conv_l^t(t))}, \\
		T_l &= \{E_i|i>=l\}\cup\{D_i|i>l\}\cup\{G\},		
	\end{aligned}
	\label{eq:conv}
\end{equation}
where $f_\uparrow$ is bilinear upsampling, $conv_l^t$ is a convolution block, and $w_l^t$ is a weight to indicate the importance of each connection.
\subsection{Searching the Optimal Connectivity}
\label{sec:loss}
The proposed FDN contains $2^{L(L+1)}$ possible final architectures, which is a huge space to search.
For example, there're $2^{30}$ architectures when the encoder has $5$ stages.
Our goal is to automatically pick the optimal connectivity out of all the possible ones.
To achieve that, we are required to (1) select the most important decoder stages from the $L$ ones, and (2) choose the input features for each selected stage.
In practice, the first problem can be reduced to the second one.
Concretely, we first select the input features for all the $L$ decoder stages, and then remove the stages without any in-connections or any out-connections, as shown in Figure~\ref{fig:framework_02} and \ref{fig:framework_03}.

Each connection in FDN contains a weight $w_l^t$ to indicate its importance, as shown in Equation~\ref{eq:conv}.
To select multiple input features out of all the possible ones for a decoder stage, the most straight forward way is to formulate $w_l^t$ as a binary indicator.
If $w_l^t=1$, the feature $t$ is chosen as the input of the $l$-th stage.
However, directly optimizing over discrete space is data inefficient, which requires lots of computation resources~\cite{liu2018darts}.
Alternatively, we propose to relax $w_l^t$ to be a continuous number between $0$ and $1$, which is then optimized by gradient descent in a differentiable manner.

\begin{figure}
\begin{center}
	\begin{subfigure}[t]{0.49\linewidth}
		\includegraphics[width=\linewidth]{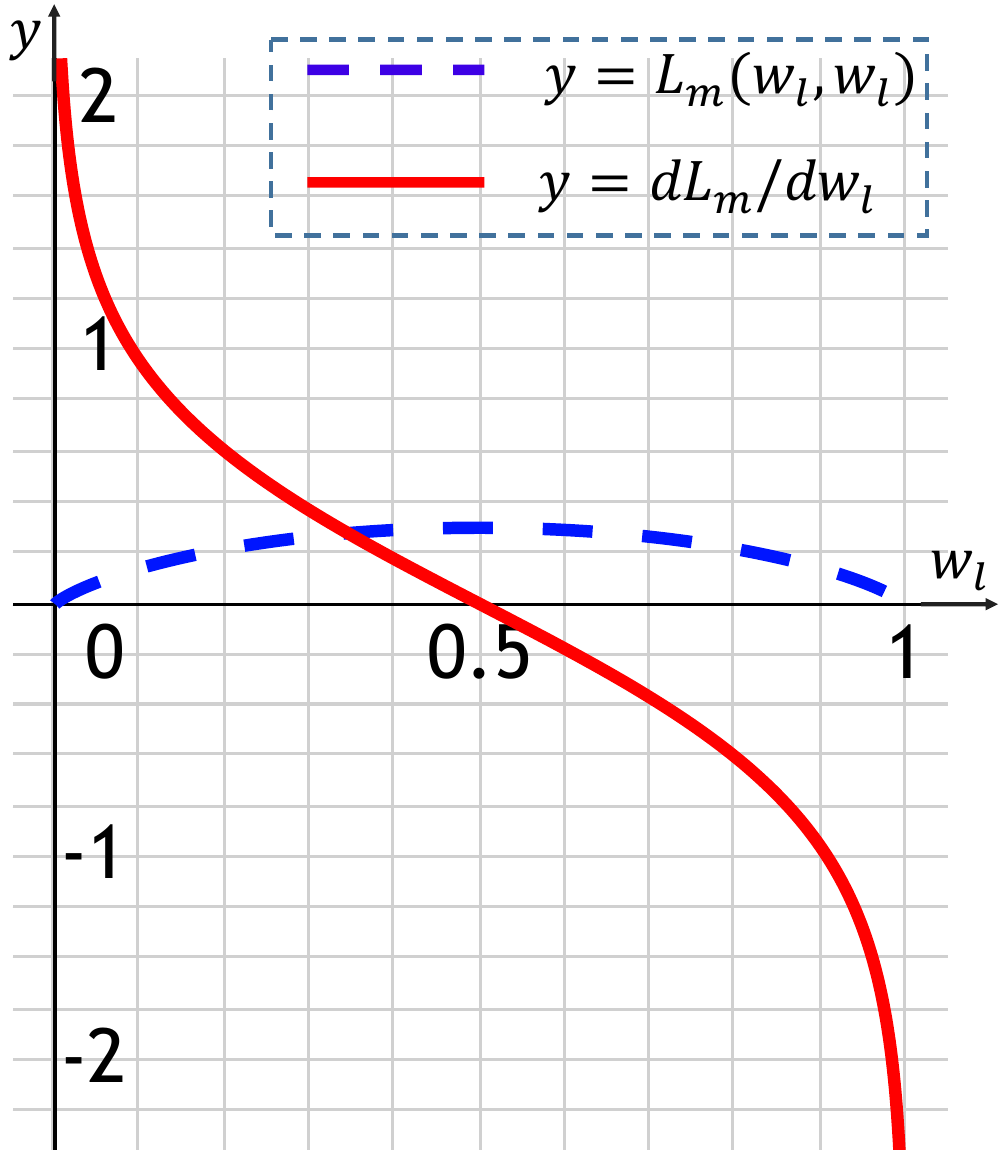}
		\caption{$L_m(w_l, w_l)$}
	\end{subfigure}
	\begin{subfigure}[t]{0.49\linewidth}
		\includegraphics[width=\linewidth]{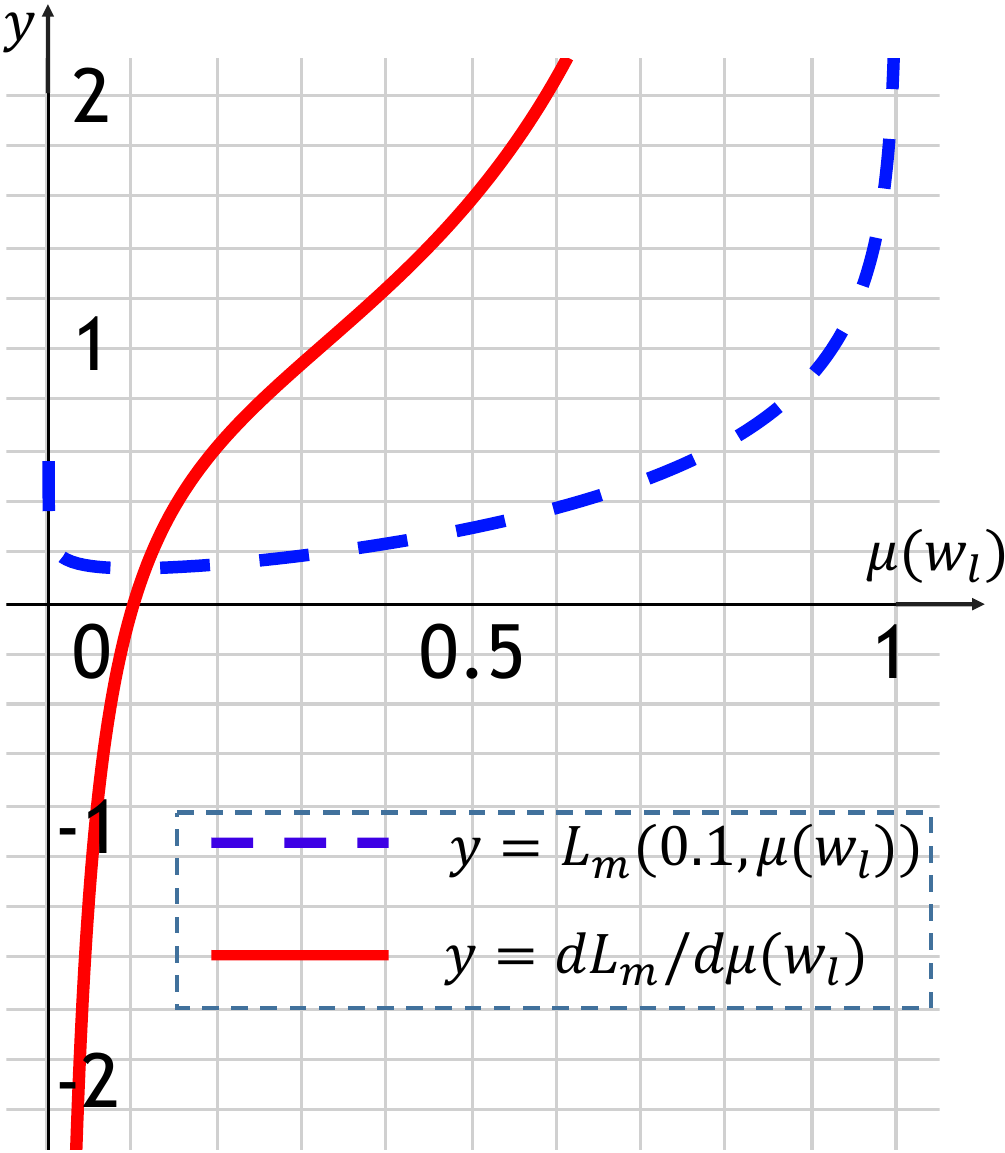}
		\caption{$L_m(\alpha, \mu(w_l))$}
	\end{subfigure}
\end{center}
   \caption{\textbf{The proposed sparse loss.} The dashed line shows the loss function while the solid line presents the gradient. (a) When minimizing $L_m(w_l, w_l)$, it forces $w_l$ close to either $0$ or $1$. (b) $L_m(\alpha, \mu(w_l))$ pushes the mean of $w_l$ near $\alpha$, which makes $w_l$ sparse when $\alpha$ is close to $0$.}
\label{fig:sparse_loss}
\end{figure}

To relax the discrete optimization problem into a continuous one, $w_l^t$ is required to satisfy a constraint that the value of $w_l^t$ needs to be close to $0$ or $1$.
Besides, $w_l=\{w_l^t|t\in T_l\}$ is desired to be sparse, since we aims at discovering an efficient architecture.
To achieve the two constraints, we propose a novel loss function in Equation~\ref{eq:sparse}, which forces $w_l^t$ to be binary and $w_l$ to be sparse.
\begin{equation}
	\begin{aligned}
		L_s(w_l, \alpha) = &~\mu(L_m(w_l, w_l)) + L_m(\alpha, \mu(w_l)), \\
		L_m(p, ~q)~     = &- p\times log(q) - (1-p)\times log(1-q), \\
    \end{aligned}
	\label{eq:sparse}
\end{equation}
where $\mu(\cdot)$ represents the mean, $\times$ is element-wise multiplication, and $\alpha$ is a hyper-parameter that controls the sparsity ratio.
As shown in Figure~\ref{fig:sparse_loss}, $L_m(w_l, w_l)$ pushes $w_l^t$ close to either $0$ or $1$, resulting in a binary-like value.
$L_m(\alpha, \mu(w_l))$ forces the mean of $w_l$ to be close to $\alpha$.
When $\alpha$ is close to $0$, most values in $w_l$ also tend to be $0$, which makes the connections sparse.
The final loss is shown in Equation~\ref{eq:loss},
\begin{equation}
	L = L_{task} + \lambda\sum_{l=1}^L L_s(w_l, \alpha),
	\label{eq:loss}
\end{equation}
where $\lambda$ controls the balance between the task-oriented loss and the sparse loss.

Notably, $w_l^t$ in Equation~\ref{eq:conv} cannot entirely indicate the importance of the connection when it is relaxed to be a continuous number.
Because the amplitude of $conv_l^t(t)$ also has an influence on the final value of $w_l^tf_\uparrow(conv_l^t(t))$.
To reduce the influence of $conv_l^t(t)$, we introduce batch normalization (BN)~\cite{ioffe2015batch} to normalize the amplitude of $conv_l^t(t)$ into $\mathcal{N}(0, 1)$:
\begin{equation}
	D_l = \sum_{t\in T_l}{w_l^tf_{\uparrow}(bn_l^t(conv_l^t(t)))}.
	\label{eq:conv_weight_bn}
\end{equation}
\subsection{Pruning Fully Dense Network}
\label{sec:prune}
To obtain the final architecture, we prune FDN according to the following rules:
(1) Drop all connections whose $w_l^t < \sigma$, where $\sigma$ is a pre-defined threshold.
(2) Drop all stages that do not have any input features.
(3) Drop all stages of which the generated feature is not used by other stages.
After pruning, $w_l^t$ and $bn_l^t$ in Equation~\ref{eq:conv_weight_bn} are removed.
The network is trained on the target dataset under the supervision of $L_{task}$ only to obtain the final result.
\section{Experiments}
In this section, we first describe the experimental details of employing our method SparseMask to automatically design a decoder for dense image prediction tasks given a pre-trained image classifier.
We then take an extensive experiment to evaluate our approach and compare the discovered architecture with other baseline methods.
\subsection{Experimental Setup}
\label{sec:4.1}
\begin{figure}
\begin{center}
	\includegraphics[width=0.95\linewidth]{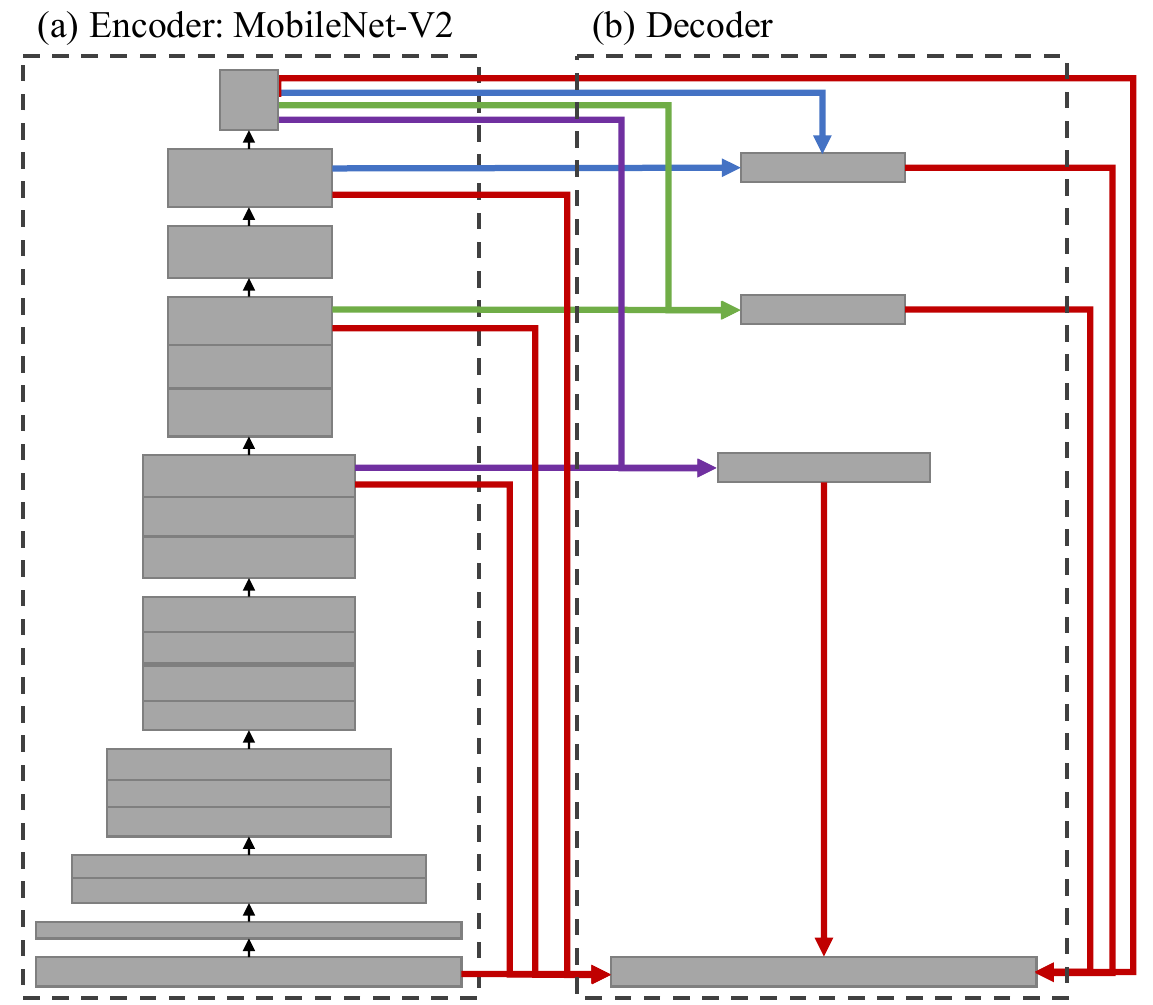}
\end{center}
    \caption{\textbf{The automatically designed architecture.} We employ SparseMask to search an efficient decoder for semantic segmentation with MobileNet-V2~\cite{sandler2018mobilenetv2} as the encoder. The training dataset is the PASCAL VOC 2012 benchmark~\cite{everingham2015pascal}. Best viewed in color.}
	\label{fig:sparse_mask}
\end{figure}
The experiment is designed to automatically search a decoder for semantic segmentation with the proposed method.
Concretely, we employ our method on the PASCAL VOC 2012 benchmark~\cite{everingham2015pascal}, which contains 20 foreground object classes and one background class.
This dataset contains $1,464$, $1,449$ and $1,456$ pixel-wise labeled images for training (\textit{train}), validation (\textit{val}), and testing (\textit{test}) respectively.
We augment the \textit{train} set with the extra annotations provided by~\cite{hariharan2011semantic}, resulting in $10,582$ training images (\textit{trainaug}).
In our experiment, the \textit{trainaug} set is used for training, while the \textit{val} set is used for testing.

To speed up the search process and reduce memory footprint, we employ MobileNet-V2~\cite{sandler2018mobilenetv2} as the pre-trained image classifier, which contains $9$ convolution stages, resulting in $2^{90}$ possible architectures\footnote{FDN based on MobileNet-V2 is shown in the supplementary material.}.
The sparsity ratio $\alpha$ in Equation~\ref{eq:sparse} is set as follows,
\begin{equation}
	\alpha_l = \min(\frac{2}{|T_l|}, 0.5), l\in[1,L],
	\label{eq:alpha}
\end{equation}
where $|T_l|$ represents the number of input features for stage $l$.
Under this setting, each decoder stage tends to select two input features.
$\lambda$ in Equation~\ref{eq:loss} is set to $0.01$.
The task-related loss $L_{task}$ is set to the pixel-wise cross entropy.

As for training, we follow the protocol presented in~\cite{chen2017rethinking}.
Concretely, we set the learning rates to $0.005$ (encoder) and $0.05$ (decoder) initially, which decrease to $0$ gradually according to the ``poly" strategy.
For data augmentation, we randomly scale (from $0.5$ to $2.0$) and left-right flip the input images, which are then cropped to $513\times 513$ and grouped with batch size $16$.
We train the network for $50$ epochs with SGD, of which the momentum is set to $0.9$ and the weight decay is set to 4e-5.
Notably, the searching is finished within 18 hours on a single Nvidia P100 GPU with $16$G memory.

After training, we prune the network with $\sigma = 0.001$ and train the final architecture following the same protocol.
The automatically designed architecture is shown in Figure~\ref{fig:sparse_mask}.
All decoder stages take the feature after global average pooling as the input, which shows its importance for semantic segmentation.
The high-level features and the low-level features are also very important, which provides the semantic information and pixel-wise location information respectively.
The middle-level information is less useful compared to other features.
\subsection{Experimental Results}
\paragraph{Performance Evaluation}
The discovered architecture is evaluated on the \textit{val} set with mean intersection-over-union (mIoU) as the metric.
As shown in Table~\ref{table:validation}, our architecture outperforms the strong baseline FCN~\cite{long2015fully} and the state-of-the-art method Deeplab-V3~\cite{chen2017rethinking} by a large margin.
Besides, our method runs much faster than Deeplab-V3 and has fewer parameters.
Notably, all the methods employ MobileNet-V2 as the backbone and follow the same training and testing protocol.
Besides, no multi-scale testing and left-right flipping are applied to the test images.
\begin{table}
\begin{center}
\begin{tabular}{l|c|c|c}
\hline
Method&mIoU&\#Params\tablefootnote{Parameters come from two parts: the encoder and the decoder.}&FPS\tablefootnote{In this paper, FPS (test phase) is measured on a Nvidia Titan-Xp GPU with a $512\times 512$ image as input.}\\
\hline
\hline
FCN~\cite{long2015fully} & 63.80\%~ & \cellcolor{First}2.22+0.03M & \cellcolor{First}156\\
Deeplab-V3~\cite{chen2017rethinking} & 72.51\%\tablefootnote{w/o COCO pre-training, multi-scale evaluation and deep supervision.} & 2.22+0.67M & ~50\\
\hline
U-Net~\cite{ronneberger2015u} & 64.72\%~ & \cellcolor{Third}2.22+0.16M & ~97\\
SparseACN~\cite{Zhu2018SparselyAC} & 72.23\%~ & 2.22+0.72M & ~74\\
\hline
FDN ($L_{task}$) & 72.72\%~ & 2.22+1.93M & ~37\\
$L_{task}+L_{sp}$ & 72.04\%~ & 2.22+1.92M & ~42\\
$L_{task}+L_{bp}$ & \cellcolor{First}73.23\%~ & 2.22+0.89M & ~76\\
Ours & \cellcolor{Third}73.18\%~ & 2.22+0.56M & \cellcolor{Third}~98\\
\hline
\end{tabular}
\end{center}
	\caption{Performance on Pascal VOC 2012 \textit{val} set with MobileNet-V2 as the backbone.}
	\label{table:validation}
\end{table}
\paragraph{Manually Designed Models} Our search space contains many classic architectures designed by experts.
To show the architecture discovered by our method is better than the others in the search space, we select two well-known architectures to compare with, namely U-Net~\cite{ronneberger2015u} and SparseACN~\cite{Zhu2018SparselyAC}.
U-Net is a classic architecture that follows the encoder-decoder style.
Compared to it, the connectivity pattern of our model is more expressive, which combines more than two input features within a decoder stage and have a better fusion of multi-scale information.
As a result, our method outperforms U-Net by a large margin (Table~\ref{table:validation}).
SparseACN proposes a pre-defined sparse connection pattern for densely connected networks, which shows a significant parameter and speed advantages without performance loss.
Compared to the pre-defined connection pattern, the connectivity of our model is more flexible and sparser.
Experiments in Table~\ref{table:validation} show that the discovered model outperforms SparseACN in mIOU, the number of parameters and FPS, which demonstrates the advantage of our approach.
We also compare with the super-network, FDN.
Results show that our method achieves a superior performance with a much sparser connectivity (Table~\ref{table:validation}).
\begin{figure}
\begin{center}
	\includegraphics[width=0.95\linewidth]{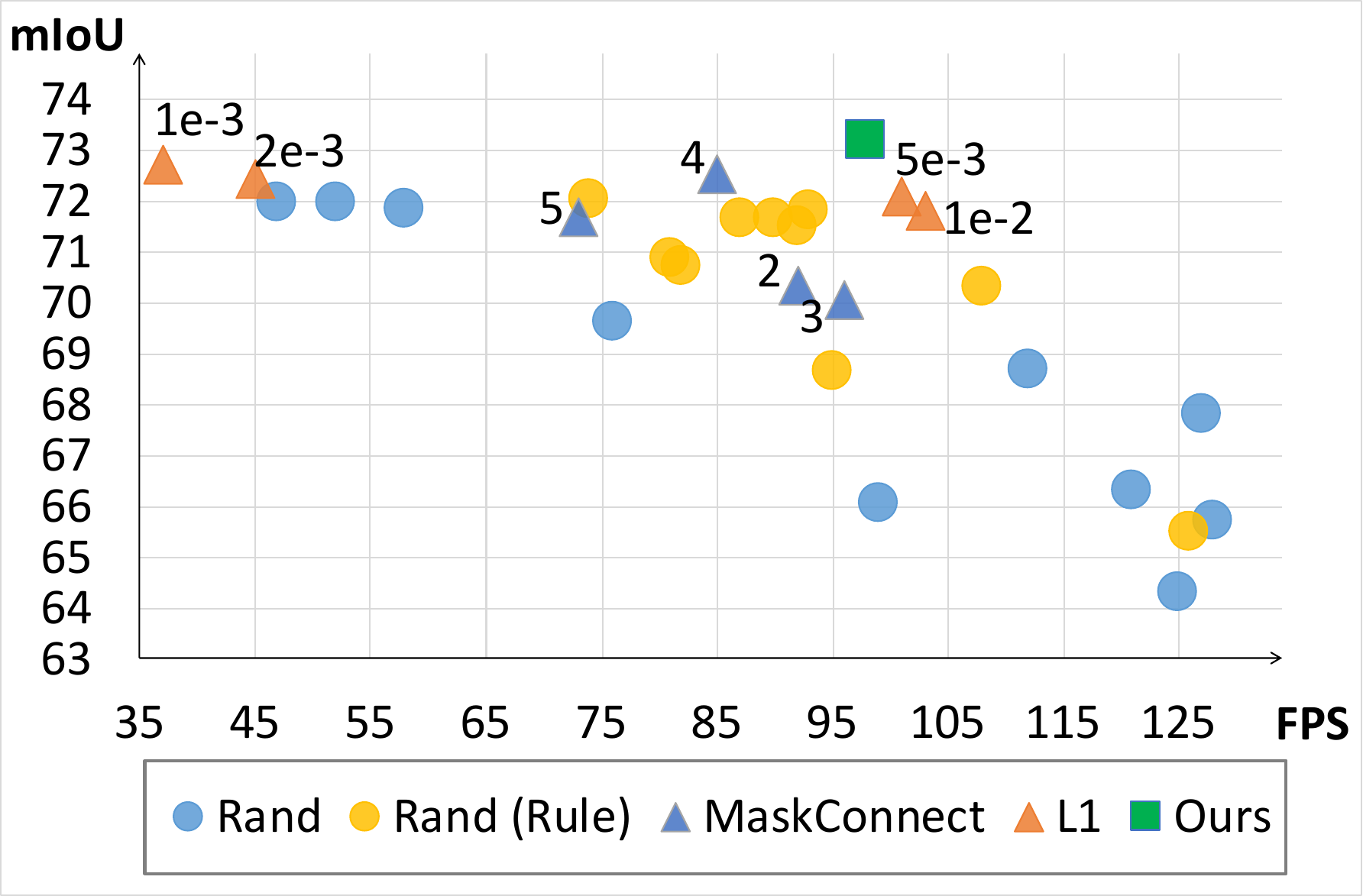}
\end{center}
	\caption{Comparison with random search, MaskConnect~\cite{ahmed2018maskconnect}, and $L_1$ loss. Best viewed in color.}
	\label{fig:random_search}
\end{figure}
\paragraph{Randomly Sampled Models}
We compare with random search to show the effectiveness of the proposed search method.
As shown in Figure~\ref{fig:random_search}, both ``Rand'' and ``Rand (Rule)'' represent the models randomly selected from our search space.
However, the models in ``Rand (Rule)'' follow a constraint observed from the connectivity pattern in Figure~\ref{fig:sparse_mask}: the input features of all but the last decoder stages are limited to $\{E_l, G\}$.
Results show that
(1) Both ``Rand'' and ``Rand (Rule)'' contain networks achieving good performance, which shows the effectiveness of our search space.
(2) Most networks in ``Rand (Rule)'' are better than that in ``Rand'', which shows that  our search method can discover a good strategy for connectivity.
(3) Our model outperforms all the randomly sampled networks, which shows the effectiveness of the proposed search method.
As for the search time, our approach is much faster than random search.
Since we only need to train the entire network once, while random search usually needs to train hundreds of sub-networks before finding a good model.
\begin{figure}
\begin{center}
	\includegraphics[width=0.9\linewidth]{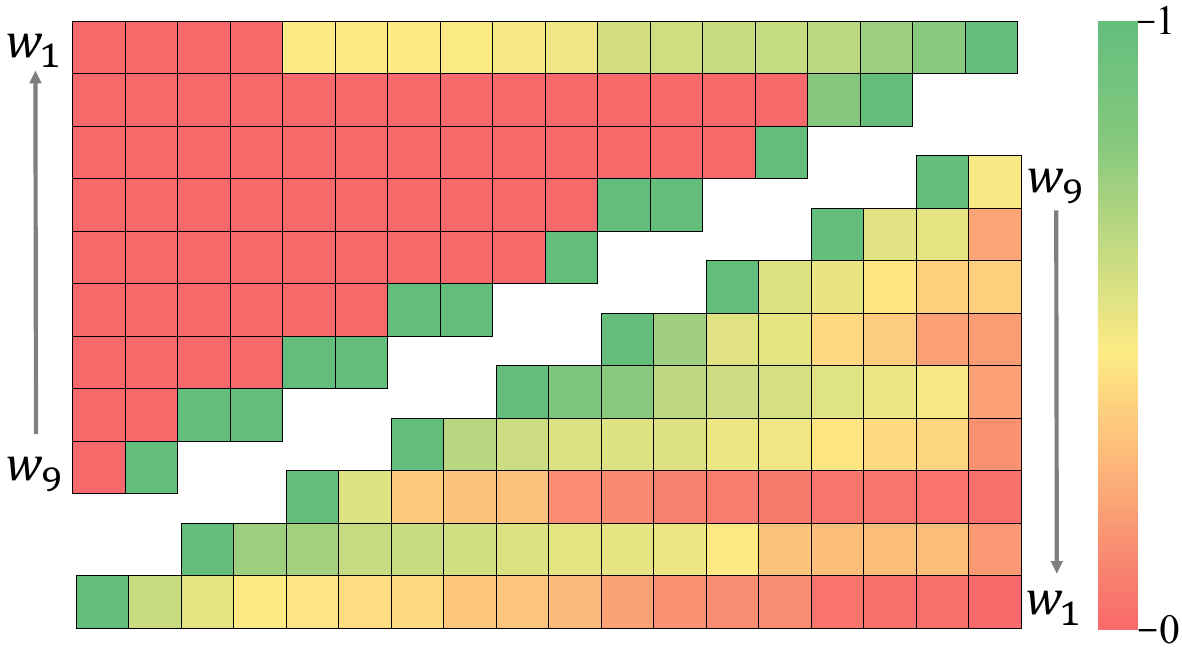}
\end{center}
   \caption{\textbf{The Sparsity of Connectivity.} Top-left is the weights trained with the proposed loss, while bottom-right is that with $L_1$ loss. The proposed loss forces $w_l$ to be binary and the connectivity to be sparse. Best viewed in color.}
\label{fig:sparsity}
\end{figure}
\paragraph{The Sparse Loss}
To show the effect of the proposed sparse loss, we take an ablation study on the binary part and the sparse part of $L_s$, noted as $L_{bp}$ and $L_{sp}$.
As shown in Table~\ref{table:validation},
(1) With $L_{task}$ only or $L_{task}+L_{sp}$, the searched connectivity is dense, resulting in low FPS and large model size.
(2) With $L_{task}+L_{bp}$, we can obtain a model with the best mIoU and a reasonable FPS, which shows the effectiveness of the binary part.
(3) By adding $L_{sp}$, the searched model (Ours) achieves similar mIoU with fewer parameters, compared to that of $L_{task}+L_{bp}$.

We also take an experiment to compare with MaskConnect~\cite{ahmed2018maskconnect} and the widely used $L_1$ loss.
As shown in Figure~\ref{fig:random_search}, we employ the search method of MaskConnect on our search space with different hyper-parameters ($k$), where $k$ is the number of selected input features for each decoder stage.
By increasing $k$ from $2$ to $5$, the searched architecture achieves a better mIoU.
However, the architecture of our method outperforms all the models discovered by MaskConnect in both mIoU and FPS.
We attribute it to the flexibility of the proposed sparse loss, which enables a much larger search space.
Concretely, our method allows an arbitrary number of input features for a decoder stage, while MaskConnect allows exactly $k$ features.
Thus, our method can discover a more expressive connection pattern in a search space with fewer constraints.

The architecture designed by $L_1$ loss also performs worse than ours, as shown in Figure~\ref{fig:random_search}.
Besides, it runs nearly $3$ times slower when the strength is 1e-3, which shows that our loss function is more effective at learning sparse connectivity.
By increasing the strength of $L_1$ loss from 1e-3 to 1e-2, the FPS improves a lot, but the performance drops significantly.
The reason is that $L_1$ loss tends to shrink all the weights, leading to the drop of useful connections after pruning.
Concretely, when the strength is 5e-3, the largest dropped connection (6.5e-4) is only 2.6$\times$ smaller than the smallest reserved one (1.7e-3), while ours is 200$\times$ smaller (1.3e-4 \vs 2.7e-2).
Thus, the influence of the dropped connections in our method can be ignored.
Such results show that our sparse loss is good at obtaining binary values.

To further show the effectiveness of our loss function, we visualize the weight of each connection.
As shown in Figure~\ref{fig:sparsity}, $w_l$ learned with our method is shown in the top-left corner, of which most squares are close to red or green.
This indicates that our method is good at approximating binary values.
Besides, the proposed loss is also helpful to sparsity, as most squares are in red.
The bottom-right corner presents the weights trained with $L_1$ loss.
The color of most squares is between red and green, which means that $L_1$ loss has little effect to approximate binary numbers.
Besides, there're only a few squares in red, which shows that $L_1$ loss has a marginal effect to sparsity.
\paragraph{Batch Normalization}
In Equation~\ref{eq:conv_weight_bn}, we employ a BN layer to normalize the feature before multiplying with $w_l^t$.
To verify its effectiveness, we conduct an experiment by removing the introduced BN layers.
As a result, the mIoU of the discovered architecture ($72.21\%$) becomes much lower than that with BN layers ($73.18\%$).
\begin{figure}
\begin{center}
	\includegraphics[width=0.9\linewidth]{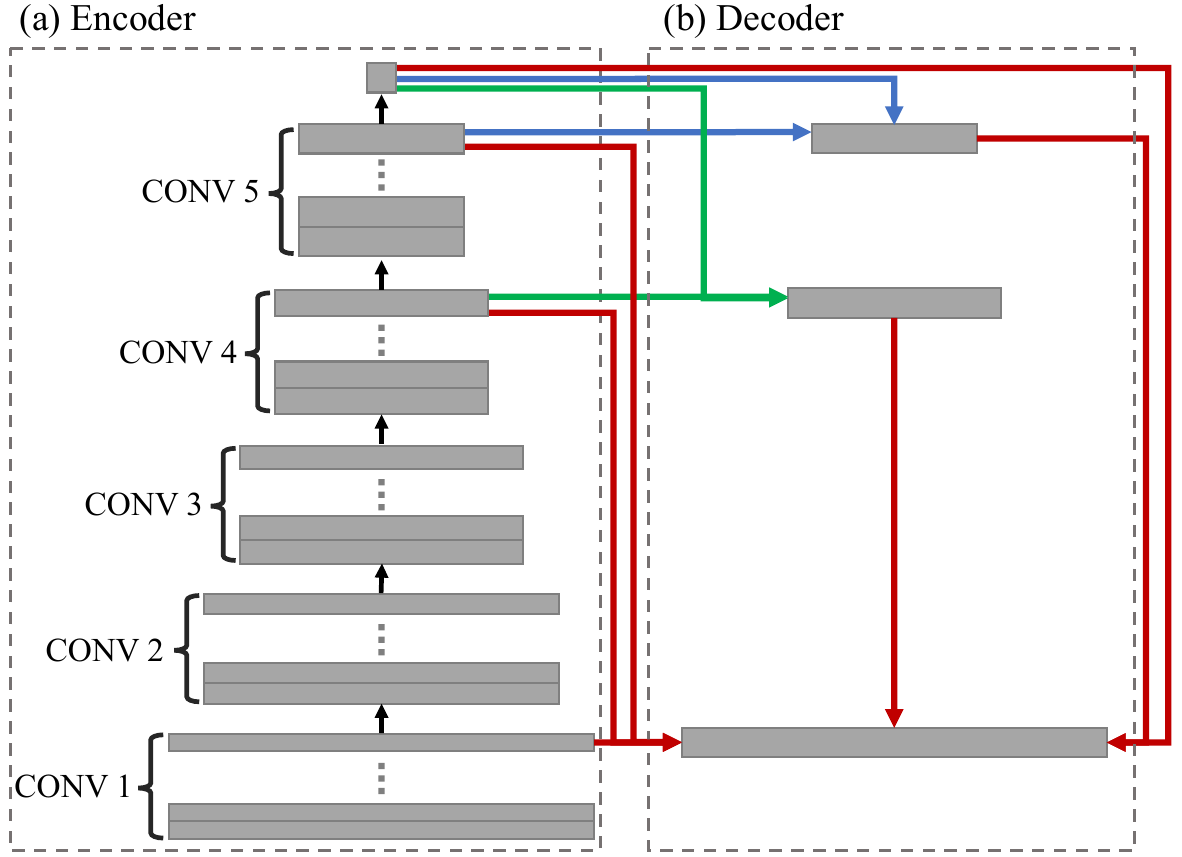}
\end{center}
   \caption{Transfer the automatically designed architecture to other backbones. The features between MobileNet-V2 and the new backbone are matched by spatial dimensions.}
\label{fig:res}
\end{figure}
\begin{table}
\begin{center}
\begin{tabular}{l|c|c|c|c}
\hline
Method&mIoU&\#Params&FPS&Memory\tablefootnote{Memory (parameters+features, training phase) are measured on a Nvidia Titan-Xp GPU with a $512\times 512$ image as the input.}\\
\hline
\hline
Deeplab-V2~\cite{chen2018deeplab}&79.7\%&-&-&-\\
RefineNet~\cite{lin2017refinenet}&84.2\%&-&-&-\\
ResNet38~\cite{wu2016wider}&84.9\%&-&-&-\\
\hline
PSPNet~\cite{zhao2017pyramid}&85.4\%&45+23M&11.5&0.9+2.3G\\
DeepLab-V3~\cite{chen2017rethinking}&85.7\%&45+16M&10.5&0.6+2.3G\\
EncNet~\cite{zhang2018context}&85.9\%&45+18M&11.7&0.8+2.3G\\
Exfuse~\cite{zhang2018exfuse}&\textbf{86.2\%}&-&-&-\\
\hline
\hline
Ours (Res101)\tablefootnote{\url{http://host.robots.ox.ac.uk:8080/anonymous/WDAEVT.html}}&85.4\%&\textbf{45+~7M}&\textbf{39.2}&\textbf{0.6+1.3G}\\
\hline
\end{tabular}
\end{center}
	\caption{Performance on the \textit{test} set of PASCAL VOC 2012 benchmark (pre-trained on MS COCO).}
	\label{table:voc_test}
\end{table}
\section{Transfer to X}
\label{sec:transfer}
The connectivity is searched for semantic segmentation on Pascal VOC 2012 benchmark with MobileNet-V2 as the backbone.
To show the generalization ability of the connectivity, we directly employ it to other backbones, datasets, and tasks without tuning.\footnote{The qualitative results are shown in the supplementary material.}
\subsection{Transfer to Other Backbones}
Our method automatically designed a connectivity for the decoder with MobileNet-V2 as the encoder.
To show the generalization ability, we transfer the discovered connectivity to other image classifiers (backbones), such as VGG nets~\cite{simonyan2014very} and ResNets~\cite{he2016deep}.
However, there're no direct correspondences between the features of different backbones.
To transfer the sparse connectivity structure, we propose to match the features by the spatial dimensions.
If a match fails, we simply drop the corresponding decoder stage.
By following the simple rule, we transfer the automatically designed connectivity to other image classifiers.
As shown in Figure~\ref{fig:res}, the encoder contains $5$ convolution stages, in which a down-sampling operation is employed followed by multiple convolution blocks.
Then a global average pooling layer is used to extract global information.
Such an encoder represents many widely used CNNs including VGG nets and ResNets.
Compared to Figure~\ref{fig:sparse_mask}, the connectivity structure is similar except that one stage is removed from the decoder because there is no corresponding feature.

To evaluate the performance of the transferred architecture, we conduct an experiment on the PASCAL VOC 2012 benchmark.
The training and testing protocol is a little different from Section~\ref{sec:4.1}.
Concretely, we have three steps for training,
(1)	train the network on MS COCO dataset~\cite{lin2014microsoft} with learning rate $0.01$ for $30$ epochs,
(2) train the network on the \textit{trainval} set of \cite{hariharan2011semantic} with learning rate $0.001$ for $50$ epochs,
and (3) train the network on the \textit{trainval} set of the original VOC dataset with learning rate $0.0001$ for $50$ epochs.
The learning rate for the decoder in each step is $10$ times larger than the above learning rate.
After training, we evaluate the model on the \textit{test} set of PASCAL VOC 2012 benchmark with multi-scale inputs and left-right flipping.

Results in Table~\ref{table:voc_test} show that our method achieves competitive performance, although the architecture is searched on MobileNet-V2 rather than ResNet101.
Moreover, our decoder requires half the parameters and runs more than $3$ times faster.
When training the network, our method occupies much less GPU memory.
Concretely, our method can be trained on a Nvidia Titan-Xp GPU with batch size $8$, while other methods like EncNet are limited to $4$ images.
\subsection{Transfer to Other Datasets}
The architecture is optimized on the Pascal VOC 2012 benchmark.
We employ it on another semantic segmentation dataset ADE20K~\cite{zhou2017scene} to verify its generalization ability.
ADE20K is a scene parsing benchmark, which contains 150 categories.
The dataset includes 20K/2K/3K images for training (\textit{train}), validation (\textit{val}) and testing (\textit{test}).

We train our network on the \textit{train} set for 120 epochs with learning rate $0.01$.
We then evaluate the model on the \textit{val} set and report the pixel-wise accuracy (PixAcc) and mIoU in Table~\ref{table:ade_20k_val}.
Our method achieves comparable results to the state-of-the-art PSPNet and EncNet, while requires much fewer parameters and runs much faster.
We then fine-tune our network on the \textit{trainval} set for another 20 epochs with learning rate $0.001$.
The outputs on the \textit{test} set are submitted to the evaluation server.
As shown in Table~\ref{table:ade_20k_test}, our method outperforms the baseline by a large margin and achieves competitive results compared to EncNet.
\begin{table}
\begin{center}
\begin{tabular}{l|cc|c}
\hline
Method&PixAcc&mIoU&Score\\
\hline
\hline
FCN~\cite{long2015fully}&71.32\%&29.39\%&50.36\%\\
SegNet~\cite{badrinarayanan2015segnet}&71.00\%&21.64\%&46.32\%\\
CascadeNet~\cite{zhou2017scene}&74.52\%&34.90\%&54.71\%\\
\hline
RefineNet~\cite{lin2017refinenet}&-&40.70\%&-\\
PSPNet~\cite{zhao2017pyramid}&81.39\%&43.29\%&62.34\%\\
EncNet~\cite{zhang2018context}&81.69\%&44.65\%&63.17\%\\
\hline
\hline
Ours (Res101)&80.91\%&43.47\%&62.19\%\\
\hline
\end{tabular}
\end{center}
	\caption{Performance on the \textit{val} set of ADE20K.}
	\label{table:ade_20k_val}
\end{table}
\begin{table}
\begin{center}
\begin{tabular}{l|cc|c}
\hline
Name&PixAcc&mIoU&Score\\
\hline
\hline
baseline-DilatedNet&65.41\%&25.92\%&45.67\%\\
rainbowsecret&71.16\%&33.95\%&52.56\%\\
WinterIsComing&-&-&55.44\%\\
CASIA\_IVA\_JD&-&-&55.47\%\\
EncNet-101~\cite{zhang2018context}&73.74\%&38.17\%&55.96\%\\
\hline
Ours (Res101)&72.99\%&38.15\%&55.57\%\\
\hline
\end{tabular}
\end{center}
	\caption{Performance on the \textit{test} set of ADE20K.}
	\label{table:ade_20k_test}
\end{table}
\subsection{Transfer to Other Tasks}
We transfer the network designed for semantic segmentation to other dense image prediction tasks, namely saliency detection~\cite{hou2017deeply} and edge detection~\cite{xie2015holistically}.
\paragraph{Saliency Detection}
MSRA-B~\cite{jiang2013salient} is a widely used dataset for saliency detection, which contains $5,000$ images with a large variation.
There are $2,500$/$500$/$2,000$ images used for training (\textit{train}), validation (\textit{val}) and testing (\textit{test}) respectively.
After training on the \textit{train} set, we evaluate our method on the \textit{test} set.
The performance is reported in Table~\ref{table:saliency}.
Our method is significantly better than the FCN baseline.
Even compared to the state-of-the-art DSS~\cite{hou2017deeply}, our method achieves comparable result with only $\frac{1}{5}$ parameters.
Besides, our method runs nearly $3$ times faster.
\begin{table}
\begin{center}
\begin{tabular}{l|cc|c|c}
\hline
Method&$F_\beta$&MAE&\#Params&FPS\\
\hline
\hline
FCN~\cite{long2015fully}&0.861&0.099&2.22+0.01M&186\\
DSS~\cite{hou2017deeply}&0.906&0.054&2.22+2.71M&~38\\
\hline
Ours&0.903&0.055&2.22+0.56M&110\\
\hline
\end{tabular}
\end{center}
	\caption{Saliency detection results on the \textit{test} set of MSRA-B. All methods are based on MobileNet-V2.}
	\label{table:saliency}
\end{table}
\paragraph{Edge Detection}
BSDS500~\cite{arbelaez2011contour} contains $200$ training (\textit{train}), $100$ validation (\textit{val}) and
$200$ test (\textit{test}) images, which is a widely used dataset in edge detection.
The \textit{trainval} set is used for training, which is augmented in the same way as \cite{xie2015holistically}.
When evaluating on the \textit{test} set, standard non-maximum suppression (NMS)~\cite{dollar2015fast} is applied to thin the detected edges.
The results are reported in Table~\ref{table:edge}, where our method outperforms the baseline method HED in two metrics (OIS and AP) and achieves the same ODS.
\begin{table}
\begin{center}
\begin{tabular}{l|cccc}
\hline
Method&ODS&OIS&AP&R50\\
\hline
\hline
HED~\cite{xie2015holistically}&\textbf{0.775}&0.792&0.826&\textbf{0.937}\\
\hline
Ours&\textbf{0.775}&\textbf{0.794}&\textbf{0.833}&0.933\\
\hline
\end{tabular}
\end{center}
	\caption{Edge detection results on the \textit{test} set of BSDS500. Both methods are based on VGG16.}
	\label{table:edge}
\end{table}
\section{Conclusion}
We presented SparseMask, a novel method that automatically designs an efficient network architecture for dense image prediction in a differentiable way, which follows the encoder-decoder style and focuses on the connectivity structure.
Concretely, we transformed an image classifier into Fully Dense Network, which contains a large set of possible final architectures and learnable dense connections.
With the supervision of the proposed sparse loss, the weight of each connection is pushed to be binary, resulting in an architecture with sparse connectivity.
Experiments show that the resulted architecture achieved competitive results on two semantic segmentation datasets, which requires much fewer parameters and runs more than $3$ times faster than the state-of-the-art methods.
Besides, the discovered connectivity is compatible with various backbones and generalizes well to many other datasets and dense image prediction tasks.
Notably, We focus on connectivity search in this paper.
However, the proposed search method can be easily extended to layer search.
We plan to combine connectivity search and layer search in our future work.
\section*{Acknowledgement}
This work is funded by the National Natural Science Foundation of China (Grand No. 61876181, 61721004, 61403383) and the Projects of Chinese Academy of Sciences (Grand QYZDB-SSW-JSC006 and Grand 173211KYSB20160008).

{\small
\bibliographystyle{ieee}
\bibliography{egbib}
}

\onecolumn
\section*{A. Theorems}
We present two theorems in Section~\ref{sec:decoder} (main text), of which the proofs are given in this section.

\begin{theorem}
Concatenating the features and then applying convolution is equal to applying convolution to each feature and then take a summation.
\end{theorem}
\begin{proof}
Given $M$ input features $F_{in}^m$ with shape $N\times C_{in}^m \times H \times W$, the concatenated feature is noted as $F_{in}$ with shape $N\times C_{in} \times H \times W$, where $C_{in} = \sum_{m=0}^{M-1}C_{in}^m$.
The corresponding convolution kernel is noted as $W$ with shape $C_{out} \times C_{in} \times \textit{KH} \times \textit{KW}$, which can be split into $M$ weights $W^m$ with shape $C_{out} \times C_{in}^m \times \textit{KH} \times \textit{KW}$.
The output feature $F_{out}$ is represented as following:
		\begin{equation}
			\begin{aligned}
				F_{out}[n, c_{out}, h, w] &= conv(F_{in}, W)[n, c_{out}, h, w] \\
				&= \sum_{kh,kw}\sum_{c_{in}=0}^{C_{in}-1}W[c_{out},c_{in},kh,kw]F_{in}[n,c_{in},h+kh,w+kw] \\
				&=\sum_{kh,kw}\sum_{m=0}^{M-1}\sum_{c_{in}=0}^{C_{in}^m-1}W^m[c_{out},c_{in},kh,kw]F_{in}^m[n,c_{in},h+kh,w+kw] \\
				&=\sum_{m=0}^{M-1}\sum_{kh,kw}\sum_{c_{in}=0}^{C_{in}^m-1}W^m[c_{out},c_{in},kh,kw]F_{in}^m[n,c_{in},h+kh,w+kw] \\
				&=\sum_{m=0}^{M-1}conv(F_{in}^m, W^m)[n, c_{out}, h, w].
		    \end{aligned}
		\end{equation}
\end{proof}
\begin{theorem}
The order of bilinear upsampling and point-wise convolution is changeable.
\end{theorem}
\begin{proof}
The input feature is $F_{in}$ with shape $N \times C_{in} \times H_{in} \times W_{in}$, while the corresponding convolution kernel is $W$ with shape $C_{out} \times C_{in} \times 1 \times 1$.
The output features $F_{out}$ is then represented as following:
		\begin{equation}
			\begin{aligned}
				F_{out}[n, c_{out}, h_{out}, w_{out}] &= conv(f_{\uparrow}(F_{in}), W)[n, c_{out}, h_{out}, w_{out}] \\
				&= \sum_{c_{in}}W[c_{out},c_{in},0,0]f_{\uparrow}(F_{in})[n,c_{in},h_{out},w_{out}] \\
				&= \sum_{c_{in}}W[c_{out},c_{in},0,0]\sum_{i=0}^3|h_{in}-h_{in}^i||w_{in}-w_{in}^i|F_{in}[n,c_{in},h_{in}^i,w_{in}^i] \\
				&= \sum_{i=0}^3\sum_{c_{in}}W[c_{out},c_{in},0,0]|h_{in}-h_{in}^i||w_{in}-w_{in}^i|F_{in}[n,c_{in},h_{in}^i,w_{in}^i] \\
				&= \sum_{i=0}^3|h_{in}-h_{in}^i||w_{in}-w_{in}^i|\sum_{c_{in}}W[c_{out},c_{in},0,0]F_{in}[n,c_{in},h_{in}^i,w_{in}^i] \\
				&= \sum_{i=0}^3|h_{in}-h_{in}^i||w_{in}-w_{in}^i|conv(F_{in}, W)[n, c_{out}, h_{in}^i, w_{in}^i] \\
				&= f_{\uparrow}(conv(F_{in}, W))[n, c_{out}, h_{out}, w_{out}],
		    \end{aligned}
		\end{equation}
where $f_\uparrow(\cdot)$ is bilinear upsampling, $h_{in} = h_{out}/H_{out}\times H_{in}$ and $w_{in} = w_{out}/W_{out}\times W_{in}$.
$h_{in}^i$ and $w_{in}^i$ is calculated as follows:
		\begin{equation}
			\begin{aligned}
				&h_{in}^0 = \lfloor h_{in} \rfloor, w_{in}^0 = \lfloor w_{in} \rfloor; h_{in}^1 = \lceil h_{in} \rceil, w_{in}^1 = \lfloor w_{in} \rfloor \\
				&h_{in}^2 = \lfloor h_{in} \rfloor, w_{in}^2 = \lceil w_{in} \rceil; h_{in}^3 = \lceil h_{in} \rceil, w_{in}^3 = \lceil w_{in} \rceil .
		    \end{aligned}
		\end{equation}
\end{proof}

\section*{B. Fully Dense Network based on MobileNet-V2}
\begin{figure*}
\begin{center}
	\includegraphics[width=0.9\linewidth]{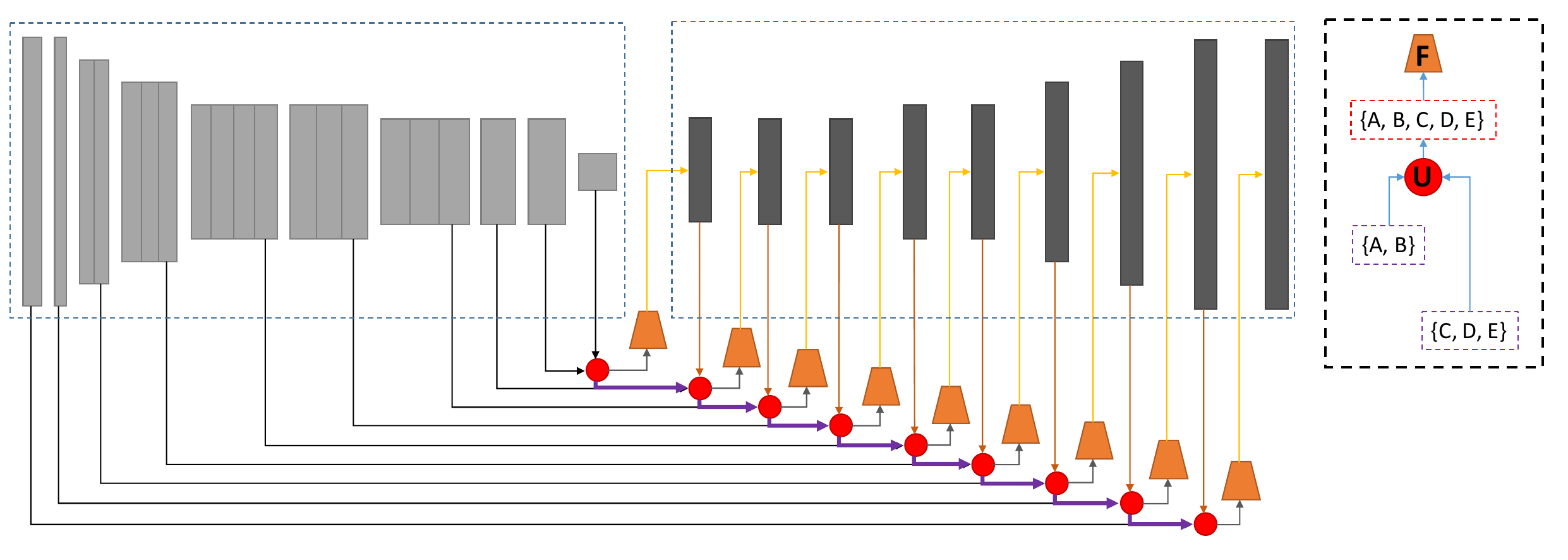}
\end{center}
	\caption{\textbf{Fully Dense Network based on MobileNet-V2~\cite{sandler2018mobilenetv2}.} The inputs to the red circle (\textbf{U}) are multiple feature sets, while the output is the union of all the sets. \textbf{F} is the decoder stage, which takes a feature set as the input. Best viewed in color.}
\label{fig:mobilenet}
\end{figure*}

Figure~\ref{fig:mobilenet} presents the Fully Dense Network based on MobileNet-V2. The inputs to the red circle (\textbf{U}) are multiple feature sets, while the output is the union of all the sets. \textbf{F} is the decoder stage, which takes a feature set as the input.

\section*{C. Visual Results}
\begin{figure*}
\captionsetup[subfigure]{labelformat=empty}
\begin{center}
	\subcaptionbox{(a) Semantic Segmentation: PASCAL VOC 2012}{
		\begin{subfigure}[b]{0.23\linewidth}
    	    \includegraphics[width=\linewidth]{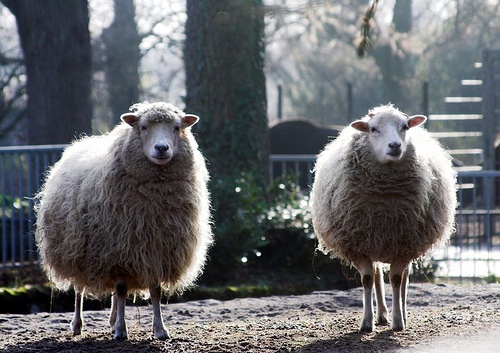}
        	\caption{Input}
	    \end{subfigure}
	    \hfill
   		\begin{subfigure}[b]{0.23\linewidth}
    	    \includegraphics[width=\linewidth]{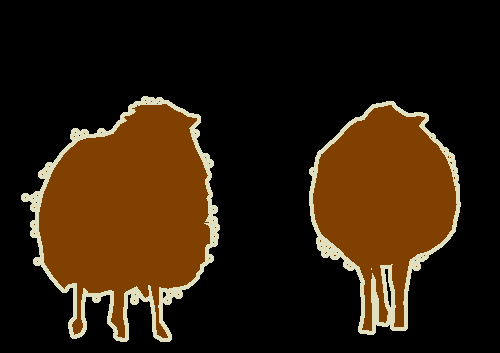}
        	\caption{GT}
	    \end{subfigure}
   	    \hfill
   		\begin{subfigure}[b]{0.23\linewidth}
    	    \includegraphics[width=\linewidth]{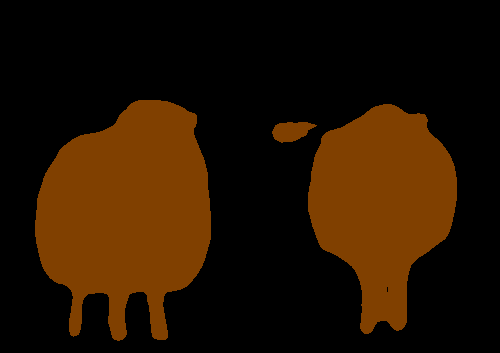}
        	\caption{EncNet~\cite{zhang2018context}}
	    \end{subfigure}
	    \hfill
   		\begin{subfigure}[b]{0.23\linewidth}
    	    \includegraphics[width=\linewidth]{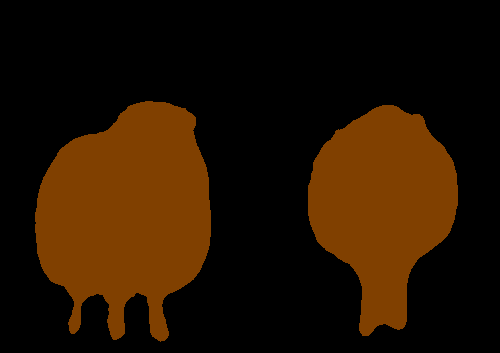}
        	\caption{Ours}
	    \end{subfigure}
	} \\
	\subcaptionbox{(b) Semantic Segmentation: ADE20K}{
		\begin{subfigure}[b]{0.23\linewidth}
    	    \includegraphics[width=\linewidth]{images/ade/input}
        	\caption{Input}
	    \end{subfigure}
	    \hfill
   		\begin{subfigure}[b]{0.23\linewidth}
    	    \includegraphics[width=\linewidth]{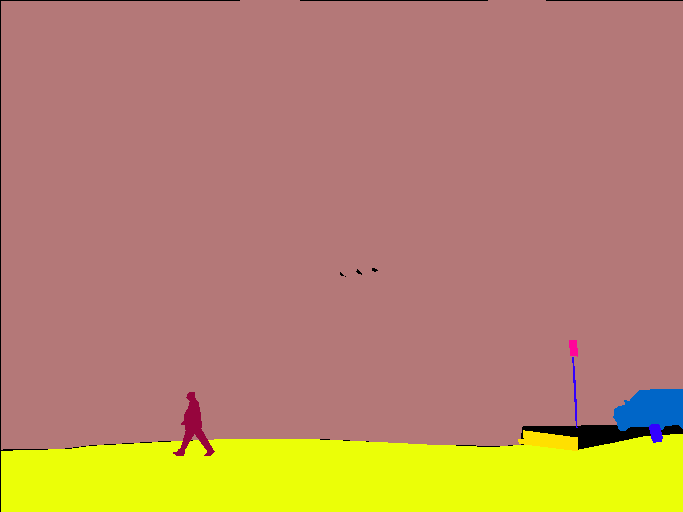}
        	\caption{GT}
	    \end{subfigure}
	    \hfill
   		\begin{subfigure}[b]{0.23\linewidth}
    	    \includegraphics[width=\linewidth]{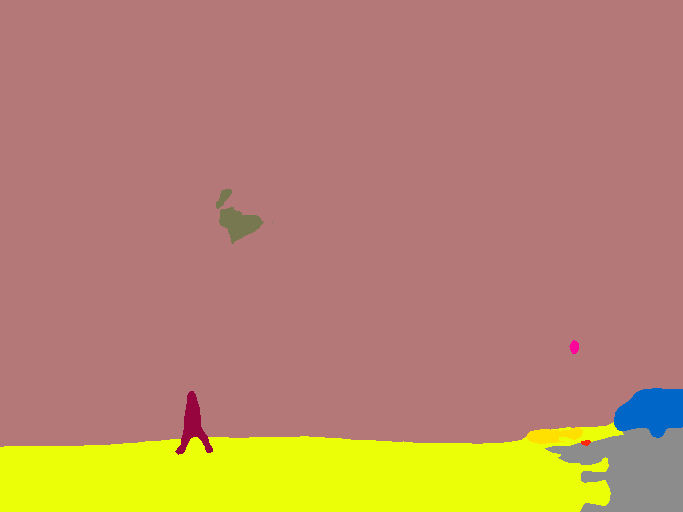}
        	\caption{EncNet~\cite{zhang2018context}}
	    \end{subfigure}
	    \hfill
   		\begin{subfigure}[b]{0.23\linewidth}
    	    \includegraphics[width=\linewidth]{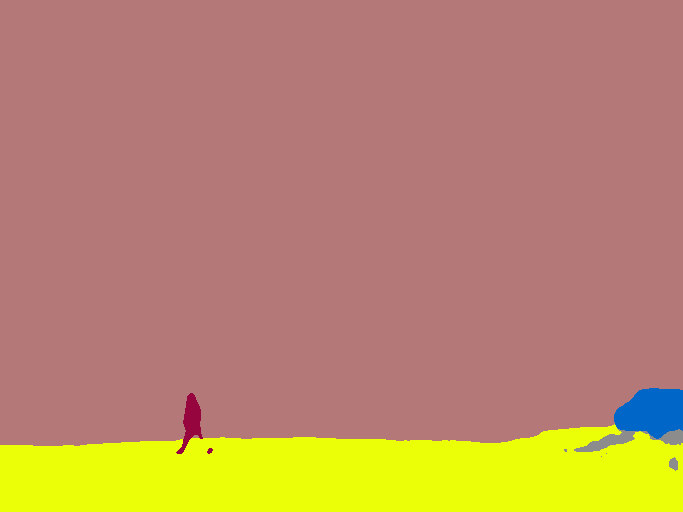}
        	\caption{Ours}
	    \end{subfigure}
	} \\
	\subcaptionbox{(c) Saliency Detection}{
		\begin{subfigure}[b]{0.23\linewidth}
    	    \includegraphics[width=\linewidth]{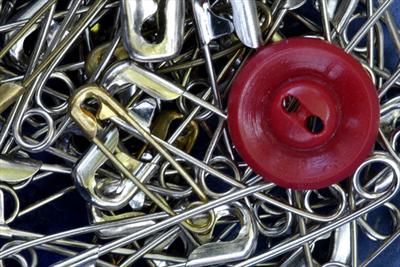}
        	\caption{Input}
	    \end{subfigure}
	    \hfill
   		\begin{subfigure}[b]{0.23\linewidth}
    	    \includegraphics[width=\linewidth]{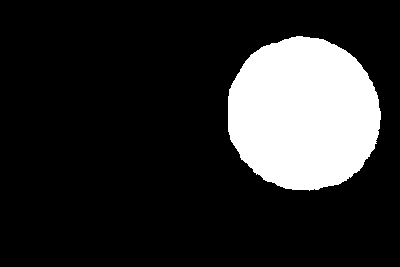}
        	\caption{GT}
	    \end{subfigure}
	    \hfill
   		\begin{subfigure}[b]{0.23\linewidth}
    	    \includegraphics[width=\linewidth]{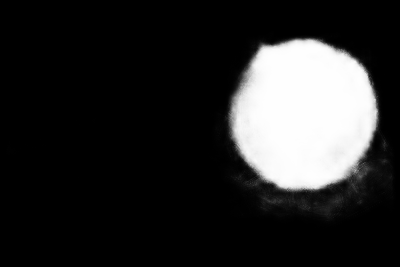}
        	\caption{DSS~\cite{hou2017deeply}}
	    \end{subfigure}
	    \hfill
   		\begin{subfigure}[b]{0.23\linewidth}
    	    \includegraphics[width=\linewidth]{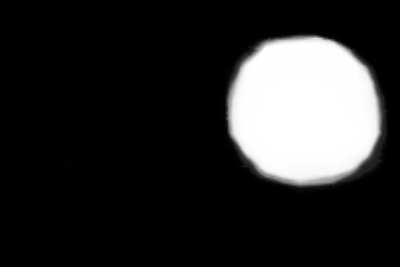}
        	\caption{Ours}
	    \end{subfigure}
	} \\
	\subcaptionbox{(d) Edge Detection}{
		\begin{subfigure}[b]{0.23\linewidth}
    	    \includegraphics[width=\linewidth]{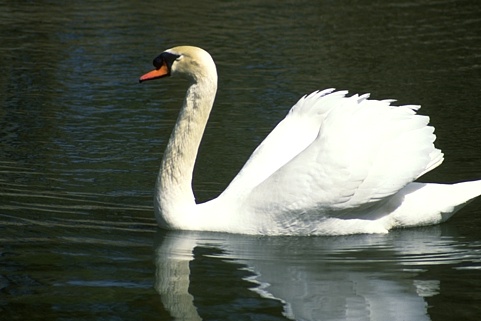}
        	\caption{Input}
	    \end{subfigure}
	    \hfill
   		\begin{subfigure}[b]{0.23\linewidth}
    	    \includegraphics[width=\linewidth]{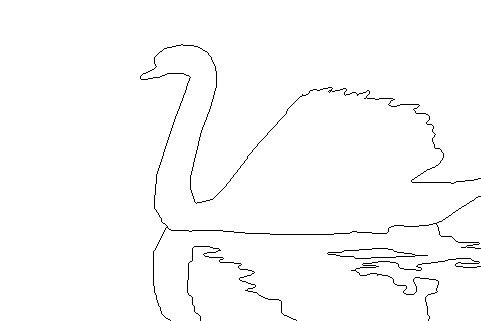}
        	\caption{GT}
	    \end{subfigure}
	    \hfill
   		\begin{subfigure}[b]{0.23\linewidth}
    	    \includegraphics[width=\linewidth]{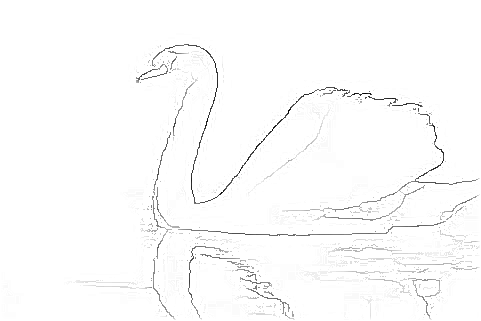}
        	\caption{HED~\cite{xie2015holistically}}
	    \end{subfigure}
	    \hfill
   		\begin{subfigure}[b]{0.23\linewidth}
    	    \includegraphics[width=\linewidth]{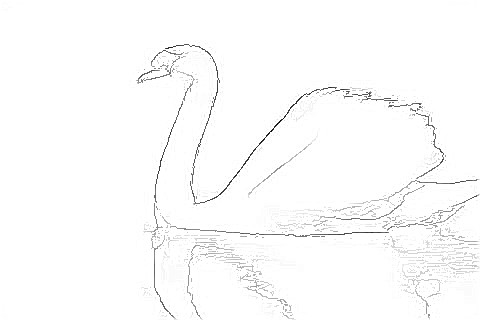}
        	\caption{Ours}
	    \end{subfigure}
	}
\end{center}
	\caption{\textbf{Qualitative Results.} Our method is not only quantitively but also qualitatively comparable to the baseline method.}
	\label{fig:visual}
\end{figure*}

The visual results for experiments in Section~\ref{sec:transfer} (main text) are shown in Figure~\ref{fig:visual}.

\end{document}